\documentclass[preprint,12pt,authoryear]{elsarticle}

\usepackage{amssymb}

\usepackage{lineno}

\usepackage{multirow} 
\usepackage{makecell} 
\usepackage{dblfloatfix} 
\usepackage{array}
\usepackage[hidelinks]{hyperref}  

\usepackage[labelsep=period]{caption}  

\journal{arXiv}

\begin{document}

\begin{frontmatter}



\title{TractoSCR: A Novel Supervised Contrastive Regression Framework for Prediction of Neurocognitive Measures Using Multi-Site Harmonized Diffusion MRI Tractography}

\author[1,2]{Tengfei~Xue}
\author[1]{Fan~Zhang\corref{cor1}}
\author[1]{Leo R. Zekelman}
\author[2]{Chaoyi~Zhang}
\author[1,2]{Yuqian~Chen}
\author[1]{Suheyla Cetin-Karayumak}
\author[1]{Steve Pieper}
\author[1]{William M. Wells}
\author[1]{Yogesh~Rathi}
\author[1]{Nikos~Makris}
\author[2]{Weidong~Cai}
\author[1]{Lauren~J.~O’Donnell\corref{cor1}}
\cortext[cor1]{Corresponding authors: fzhang@bwh.harvard.edu (Fan Zhang) and odonnell@bwh.harvard.edu (Lauren J. O’Donnell).}

\address[1]{Brigham and Women’s Hospital, Harvard Medical School, Boston, USA}
\address[2]{School of Computer Science, University of Sydney, Sydney, Australia}



\begin{abstract}
Neuroimaging-based prediction of neurocognitive measures is valuable for studying how the brain’s structure relates to cognitive function. However, the accuracy of prediction using popular linear regression models is relatively low. We propose a novel deep regression method, namely \textit{TractoSCR}, that allows full supervision for contrastive learning in regression tasks using diffusion MRI tractography. TractoSCR performs supervised contrastive learning by using the absolute difference between continuous regression labels (i.e. neurocognitive scores) to determine positive and negative pairs. We apply TractoSCR to analyze a large-scale dataset including multi-site harmonized diffusion MRI and neurocognitive data from 8735 participants in the Adolescent Brain Cognitive Development (ABCD) Study. We extract white matter microstructural measures using a fine parcellation of white matter tractography into fiber clusters. Using these measures, we predict three scores related to domains of higher-order cognition (general cognitive ability, executive function, and learning/memory). To identify important fiber clusters for prediction of these neurocognitive scores, we propose a permutation feature importance method for high-dimensional data. We find that TractoSCR improves the accuracy of neurocognitive score prediction compared to other state-of-the-art methods. We find that the most predictive fiber clusters are predominantly located within the superficial white matter and projection tracts, particularly the superficial frontal white matter and striato-frontal connections. Overall, our results demonstrate the utility of contrastive representation learning methods for regression, and in particular for improving neuroimaging-based prediction of higher-order cognitive abilities.
\end{abstract}



\begin{keyword}


Diffusion MRI \sep Neurocognition prediction \sep ABCD study \sep Deep learning \sep Contrastive representation learning
\end{keyword}

\end{frontmatter}


\section{Introduction}
\label{sec:introduction}
The brain’s white matter (WM) connections, which can be quantitatively mapped using diffusion MRI (dMRI) tractography \citep{Zhang2022-sj}, play an important role in brain networks that enable human cognition \citep{Wang2018-kd,Zekelman2022-lz}. Investigating the predictive relationship between WM microstructure and cognition can therefore improve our understanding of the brain in health and disease. Regression analysis, which can predict values of a dependent variable (label) given a set of input independent variables (features), enables the prediction of neurocognitive measures given input features from neuroimaging. This strategy is recently of high interest~\citep{Sripada2020-pg, Kim2021-kp, Wu2022-rn, Radhakrishnan2022-fc,Feng2022-xf}. While many studies perform prediction using high-dimensional neuroimaging features from functional MRI (fMRI) \citep{Cui2018-uu,Dubois2018-zk,Sripada2020-pg,Wu2022-rn} or multimodal data \citep{Gong2021-up,Gong2022-wl,Mansour_L2021-xs,Kim2021-kp,Sun2022-qu,Radhakrishnan2022-fc}, a unimodal focus on dMRI tractography (e.g.,~\citep{Jeong2021-yv, Feng2022-xf, Mansour_L2022-vg, Chen2022-nl}) can improve our understanding of the role of the WM connections in cognition. While a number of studies have pursued prediction of neurocognitive measures based on information from dMRI tractography (Table~\ref{tab_related_works}), current approaches are limited in terms of study cohorts and regression methodology.

\newcolumntype{M}[1]{>{\centering\arraybackslash}m{#1}} 
\begin{table}[hbtp]
\caption{Summary of studies for prediction of neurocognitive measures using white matter measures extracted only from dMRI tractography. For each study, study cohorts and regression methodology are reported.}
\setlength{\tabcolsep}{1.5pt} 
\begin{tabular}{|M{90pt}| M{150pt}| M{140pt}|}
\hline
\textbf{Study cohorts} & Number of subjects $\leqslant$ 1000 &Number of subjects $>$ 1000  \\
\hline
Neonates & \citep{Chen2020-cs},~\citep{Chen2022-nl} & None \\
\hline
Children & \citep{Jeong2021-yv} & \textbf{Our current study} \\
\hline
Young adults & \citep{noauthor_undated-tc},~\citep{Jandric2022-ki},~\citep{Seguin2020-hr},~\citep{Zekelman2022-lz} & None \\
\hline
Older adults & \citep{Li2020-le},~\citep{Aben2021-lz},~\citep{OSullivan2022-zo},~\citep{Jandric2022-ki},~\citep{Berger2022-hq} & \citep{Feng2022-xf},~\citep{Madole2021-io} \\
\hline
\multicolumn{3}{|p{380pt}|}{\textbf{Regression methodology}} \\
\hline
Linear model & \multicolumn{2}{M{290pt}|}{\citep{noauthor_undated-tc},~\citep{Feng2022-xf},~\citep{Madole2021-io},~\citep{Li2020-le},~\citep{Jandric2022-ki},~\citep{Seguin2020-hr},~\citep{Zekelman2022-lz},~\citep{Berger2022-hq}} \\
\hline
Non-linear SVM & \multicolumn{2}{M{290pt}|}{\citep{OSullivan2022-zo}} \\
\hline
MLP & \multicolumn{2}{M{290pt}|}{\citep{Feng2022-xf}} \\
\hline
CNN & \multicolumn{2}{M{290pt}|}{\citep{Jeong2021-yv},~\citep{Chen2020-cs},~\citep{Chen2022-nl}} \\
\hline
Representation learning & \multicolumn{2}{M{290pt}|}{\textbf{Our current study}} \\
\hline
\multicolumn{3}{p{370pt}}{\textit{Abbreviations}: SVM, support vector machine; MLP, multilayer perceptron; CNN, convolutional neural network.}\\
\end{tabular}
\label{tab_related_works}
\end{table}

Linear regression models such as ElasticNet \citep{Zou2005-oq} have been widely used for prediction of neurocognitive performance \citep{Cui2018-uu,Jollans2019-ma,Sripada2020-pg,Gong2021-up,noauthor_undated-tc,Feng2022-xf,Madole2021-io,Li2020-le,Jandric2022-ki,Seguin2020-hr,Zekelman2022-lz}, while some studies \citep{Feng2022-xf,Chen2022-nl,Jeong2021-yv} have explored deep-learning-based regression using multilayer perceptrons (MLP) and convolutional neural networks (CNN). However, the prediction accuracy of linear regression models is relatively low \citep{Sripada2020-pg}, and non-linear regression models may suffer from overfitting, especially on high-dimensional datasets \citep{Cui2018-uu}. Developing more advanced methods has the potential to improve prediction accuracy of neurocognitive performance metrics and to provide novel information about specific brain structures that may be important for their prediction.

One avenue for improving the prediction of neurocognitive performance metrics is to investigate recent machine learning algorithms for the analysis of tabular (row and column) data \citep{Borisov2021-kz}. Many quantitative features derived from neuroimaging can be represented as tabular data. The most popular machine learning algorithm for tabular data is the gradient boosting decision tree (GBDT) method \citep{Chen2016-em,Prokhorenkova2018-mr}. In recent years, deep-learning-based methods \citep{Yoon2020-xu,Arik2021-cc,Gorishniy2021-ll,Bahri2021-li} have been developed for tabular data, which is the last “unconquered castle” for deep learning \citep{Kadra2021-rz,Borisov2021-kz}. One important research direction for deep learning on tabular data is representation learning, which can discover beneficial data representations for downstream tasks. For example, the value imputation and mask estimation (VIME) \citep{Yoon2020-xu} and self-supervised contrastive learning using random feature corruption (SCARF) \citep{Bahri2021-li} methods enable representation learning on tabular data. However, these representation learning methods were developed for classification tasks, and cannot utilize regression label information during representation learning. 

Another avenue for improving prediction of neurocognitive measures is to investigate recently proposed algorithms for contrastive learning \citep{Chen2020-pz,Chen2021-il,Khosla2020-mx}. In medical image computing, supervised contrastive learning improves classification accuracy by using labels during representation learning \citep{Schiffer2021-vl,Dufumier2021-bk,Xue2022-yz,Seyfioglu2022-bj}. It is usually designed for classification tasks, where samples with the same categorical label are positive pairs, and samples with different categorical labels are negative pairs. During representation learning, embeddings of positive pairs are pulled together, and embeddings of negative pairs are pushed apart. However, regression tasks require continuous labels (e.g. neurocognitive scores) that cannot directly be used for pair determination. Two recent works have shown that contrastive learning can be useful in the context of regression based on medical images as input \citep{Lei2021-ey, Dai2022-xu}. For example, RPR-Loc proposed a learning strategy to predict the distance between a pair of image patches \citep{Lei2021-ey}. Recently, the AdaCon method used a contrastive learning strategy that leveraged distances between labels (e.g. bone mineral densities) to benefit downstream computer-aided disease assessment. These recent regression methods did not use labels for pair determination for contrastive learning. How to best use label information to enhance regression is still an open question. 

\begin{figure*}[ht]
\centering
\includegraphics[width=\textwidth]{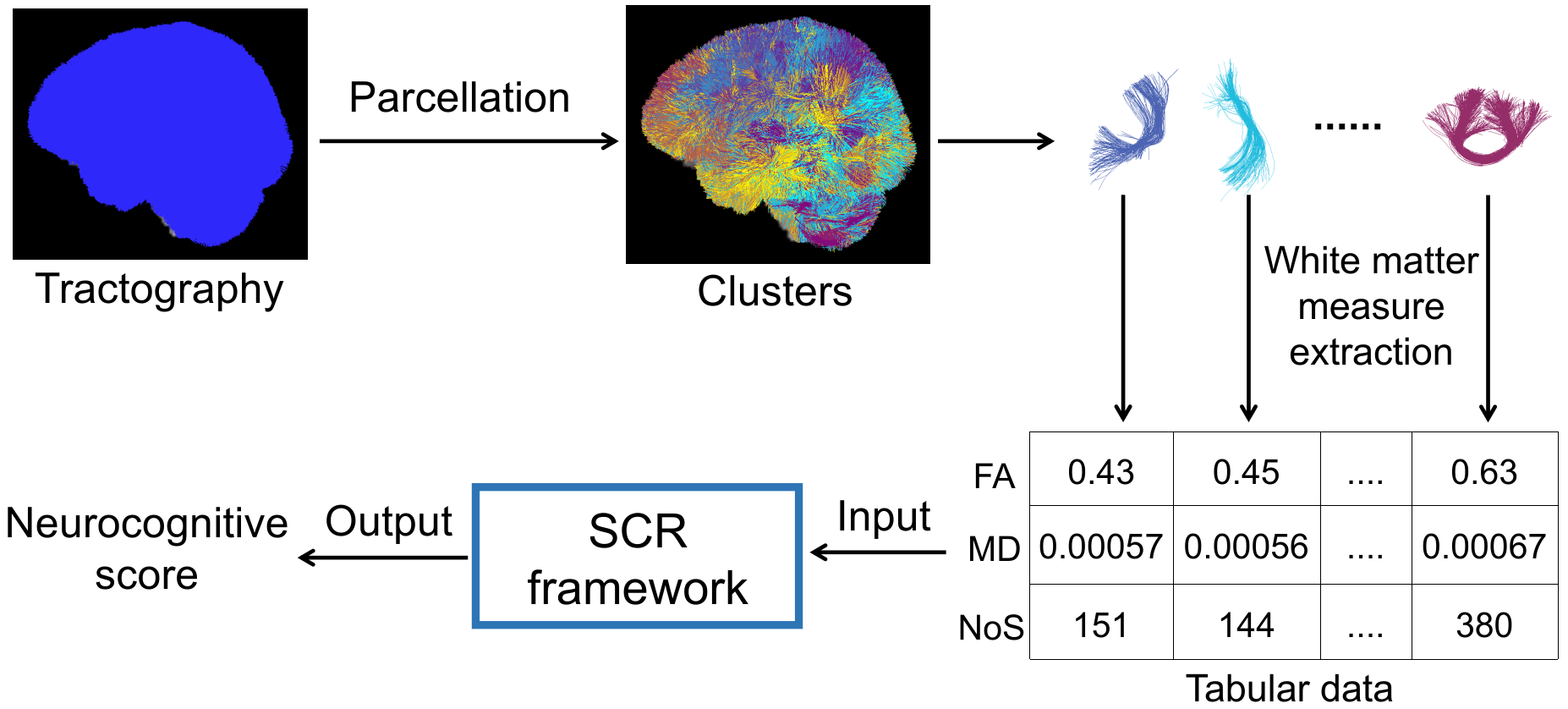}
\caption{Overview of our proposed TractoSCR framework for neurocognitive score prediction using dMRI tractography. Parcellation of tractography into fiber clusters enables the extraction of cluster-specific white matter measures. These measures are represented as tabular data and input to the TractoSCR framework, which outputs a neurocognitive score. FA: fractional anisotropy, MD: mean diffusivity, NoS: number of streamlines.}
\label{fig_framework_overview}
\end{figure*}

In this study, we propose a novel deep regression method for tractography analysis with supervised contrastive regression, referred to as \textit{TractoSCR}. TractoSCR is a novel contrastive representation learning framework to predict measures of neurocognition using white matter microstructure derived from dMRI tractography, as illustrated in Fig.~\ref{fig_framework_overview}. Our proposed TractoSCR method extends the supervised contrastive learning method \citep{Khosla2020-mx}, which is designed for categorical data in classification tasks, to perform regression analysis where the predicted labels are continuous values. We propose a novel pair-determination strategy that uses the absolute difference between continuous regression labels to determine positive and negative sample pairs for contrastive learning. To our knowledge, this is the first method that leverages deep representation learning techniques for the prediction of neurocognitive performance. Our method uses a tractography fiber clustering method that enables consistent white matter parcellation across populations. The parcellation allows representation of microstructure features from whole brain tractography as tabular data, which enables the use of a recently proposed random feature corruption technique \citep{Bahri2021-li} for data augmentation to further improve prediction performance. In addition, for interpreting prediction results, we propose a novel permutation feature importance algorithm to identify tractography fiber clusters and their corresponding anatomical tracts that are important for prediction of neurocognitive measures. We demonstrate our method in a large-scale dMRI dataset including data from 8735 children, where we explore the relationship between white matter microstructure and prediction of neurocognitive performance (including general ability, executive function, and learning/memory). 

The remaining structure of this paper is as follows. Section~\ref{sec_methods} describes the dataset and data processing, the proposed regression and interpretation methods, and the model training and testing details. Section~\ref{sec_experiments} describes the evaluation metric, experimental results, and interpretation of results. Finally, the discussion and conclusion are given in Section~\ref{sec_discussion} and~\ref{sec_conclusion}, respectively.

\section{Methodology}
\label{sec_methods}
\subsection{ABCD Dataset, Tractography Parcellation, and Microstructural Measures}

This study includes dMRI data and neurocognitive component scores from the Adolescent Brain Cognitive Development (ABCD) dataset\footnote{Download at https://nda.nih.gov/abcd} for 8735 American children (4560 males and 4175 females) between the ages of 9-11 (9.9±0.6) across 21 data collection sites \citep{Casey2018-xx,Volkow2018-og}. Three neurocognitive principal component scores from ABCD were studied, representing three major domains of higher-order cognition, namely \textit{General Ability} (PC1), \textit{Executive Function} (PC2) and \textit{Learning/Memory} (PC3) \citep{Thompson2019-jt}. These component scores are lower dimensional representations of nine assessment measures from the ABCD neurocognitive battery \citep{Luciana2018-oj} (including seven measures from the NIH toolbox \citep{Casaletto2015-vi}). These component scores statistically summarize nine neurocognitive assessment measures and reveal latent variables which have been theorized to be a more pure reflection of the cognitive domains of interest \citep{Snyder2015-zt,Thompson2019-jt}. Furthermore, these component scores have been associated with measures of psychopathological behavior (i.e., stress reactivity and/or externalizing behaviors), perhaps suggesting their clinical utility \citep{Thompson2019-jt}.  

The ABCD dMRI data was harmonized \citep{Cetin_Karayumak2019-dd,Zhang2022-dm,Cetin-Karayumak2022-mq,Suheyla_Cetin-Karayumak_Fan_Zhang_Tashrif_Billah_Sylvain_Bouix_Steve_Pieper_Lauren_J_ODonnell_and_Yogesh_Rathi2021-bi} to remove scanner-specific biases, allowing for a large-scale data-driven way to study relationships between brain microstructure and neurocognition. The dMRI harmonization method \citep{Cetin_Karayumak2019-dd} retrospectively removes scanner-specific differences from raw dMRI signals across disparate sites and acquisition parameters, while preserving inter-subject biological variability (e.g., fractional anisotropy (FA) values) \citep{Zhang2022-dm}. 

A two-tensor Unscented Kalman Filter (UKF) tractography method\footnote{https://github.com/pnlbwh/ukftractography} \citep{Malcolm2010-mk,Reddy2016-ko} was conducted on harmonized dMRI data of all subjects to obtain whole-brain tractography. The UKF method fitted a mixture model of two tensors to the diffusion data while tracking streamlines. This enabled the estimation of fiber-specific microstructural measures from the first tensor, which models the tract being traced \citep{Reddy2016-ko}. Next, automated parcellation of tractography was performed based on an anatomically curated cluster atlas\footnote{https://github.com/SlicerDMRI/ORG-Atlases}~\citep{Zhang2018-jx}, which was provided by the O’Donnell Research Group (ORG). Compared to traditional tractography parcellation based on cortical atlases, this clustering method was shown to be more reproducible and consistent across the lifespan \citep{Zhang2018-jx,Zhang2019-bl}. For each subject, the ORG atlas \citep{Zhang2018-jx} enabled extraction of 953 expert-curated fiber clusters. These finely parcellated fiber clusters are grouped and categorized into 58 deep white matter tracts including major long range association and projection tracts, commissural tracts, and tracts related to the brainstem and cerebellar connections, as well as 198 short and medium range superficial fiber clusters. We performed tractography quality control and white matter parcellation using open-source WhiteMatterAnalysis (WMA)\footnote{https://github.com/SlicerDMRI/whitematteranalysis} software. Tractography visualization was performed using SlicerDMRI software\footnote{dmri.slicer.org} \citep{Norton2017-nz,Zhang2020-cv}. 

For all subjects, cluster-specific microstructural measures of fractional anisotropy (FA), mean diffusivity (MD), and number of streamlines (NoS) were computed. These measures have been previously shown to be associated with neurocognitive scores \citep{Zekelman2022-lz,Chen2022-wq,Madole2021-io}. Here, FA and MD are measures of fiber-specific tissue microstructure, while NoS is widely used to quantify the connectivity strength \citep{Zhang2022-sj}. These cluster-specific measures can be considered as tabular data, allowing algorithms from the field of tabular data to be employed. For any empty cluster (due to variability of tractography or the underlying anatomy), each measure was set to zero, as in \citep{He2022-po}.

\begin{figure*}[t]
\centering
\includegraphics[width=1.0\textwidth]{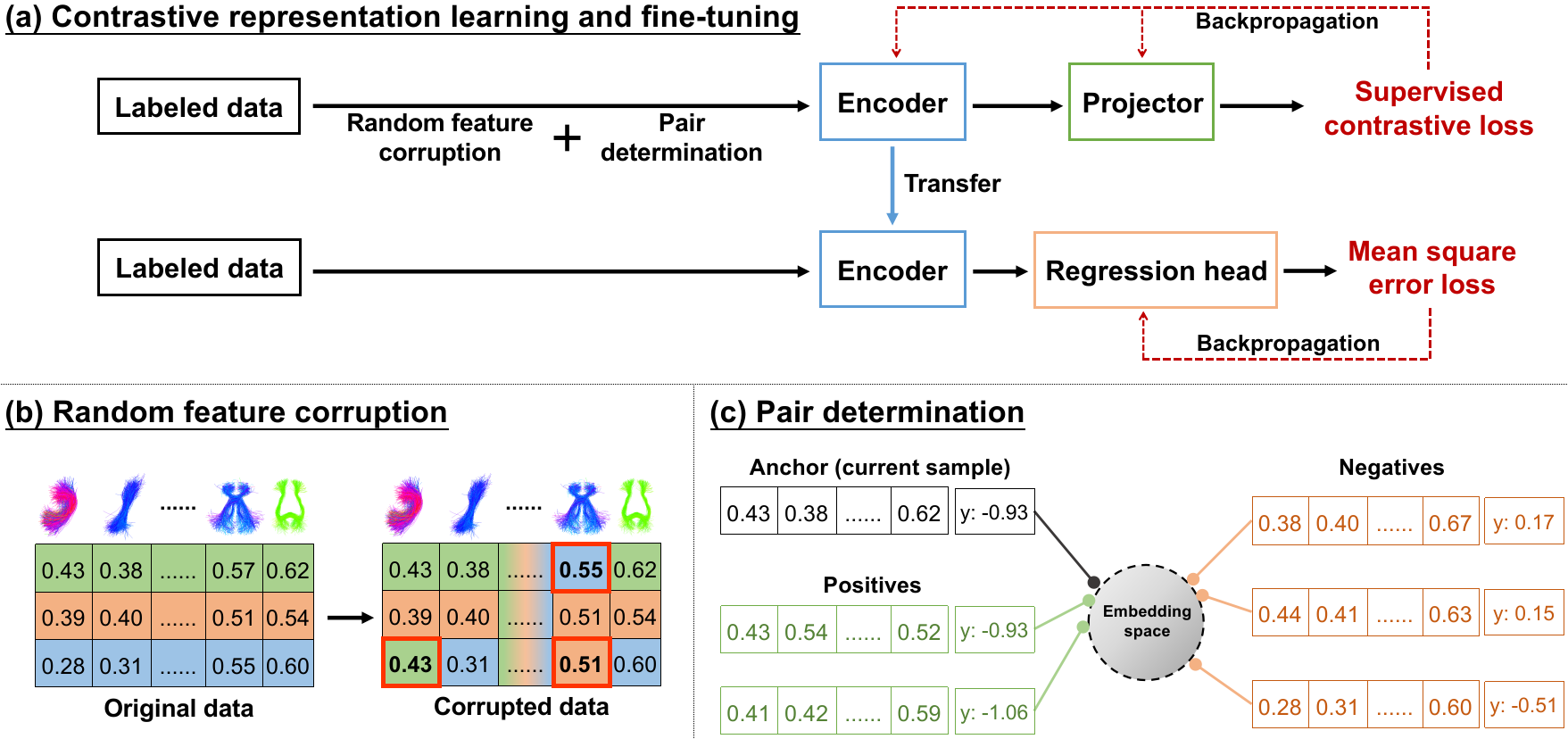}
\caption{TractoSCR framework: (a) overview of contrastive representation learning and fine-tuning, (b) random feature corruption for data augmentation with a measure of interest (e.g., FA) (rows are randomly selected samples, and columns are cluster-specific microstructural measures), (c) positive and negative pairs determination with regression labels (e.g., PC1).}
\label{fig_scr}
\end{figure*}

\subsection{Supervised Contrastive Regression}
We propose a novel contrastive representation learning method for regression, \textit{TractoSCR}. Our overall strategy is to use the absolute difference between two continuous regression labels to determine positive and negative pairs for contrastive learning. An overview of the TractoSCR framework is shown in Fig.~\ref{fig_scr}. The regression framework (Fig.~\ref{fig_scr}~(a)) has two phases: contrastive representation learning and fine-tuning. In representation learning, random feature corruption (Fig.~\ref{fig_scr}~(b)) and proposed pair determination (Fig.~\ref{fig_scr}~(c)) are utilized with a supervised contrastive loss. The network trained in representation learning is then fine-tuned to output neurocognitive scores. These steps are described in the following sections.

\subsubsection{Random Feature Corruption for Data Augmentation}
To avoid potential model overfitting and increase the discriminative ability of the learned global features in contrastive learning, we performed a data augmentation process to create more training samples. We applied the recently proposed random feature corruption technique that was designed specifically for tabular data \citep{Yoon2020-xu,Bahri2021-li} . In brief, in each mini-batch of training with input samples $X$, we created a corrupted batch copy $\tilde{X}$. To do so, we chose a proportion of the input cluster-specific measures (features) uniformly at random and replaced each of those measures by a random draw from the corresponding measure dimension of other samples (as shown in Fig.~\ref{fig_scr}~(b)). The ratio of replaced measures to all measures is defined as the corruption rate $c$. Corrupted samples $\tilde{X}$ retain the same regression labels $Y$ as original samples $X$.

\subsubsection{Positive and Negative Pairs Determination}
From the generated augmented data in each training mini-batch, we construct positive and negative sample pairs to enable supervised contrastive learning (SCL). Unlike existing studies \citep{Khosla2020-mx} using SCL to perform a classification task, where positive and negative pairs are defined based on the class labels, determination of positive and negative sample pairs is not straightforward in regression because the regression labels are continuous values. To handle this, we propose a new strategy that uses the absolute difference between two continuous regression labels to determine pairs (Fig.~\ref{fig_scr}~(c)). Given $x_{i}, x_{j}\in X$ with labels $y_{i}$ and $y_{j}$, if $|y_{i}-y_{j}| < \theta$, $x_{i}$ and $x_{j}$ are defined as positive pairs. Otherwise, $x_{i}$ and $x_{j}$ are considered to be negative pairs. The label difference threshold $\theta$, a threshold on the absolute difference of two regression labels, is the key parameter for positive and negative pair determination. For our dataset with regression labels ranging from approximately -3 to 3, the optimal $\theta$ is 0.35 based on experimental results. Note that our TractoSCR method is robust to changes in this threshold (from 0.1 to 0.5) as described in Section~\ref{subsubsec_hyperparameter}.

\subsubsection{Supervised Contrastive Loss}
After positive and negative pairs are determined using regression labels, the supervised contrastive loss as shown below becomes applicable:
$$\mathcal{L}=\sum_{r \in R}\mathcal{L}_{r}=\sum_{r \in R} \frac{-1}{|P(r)|}  \sum_{p \in P(r)} \log \frac{\exp ( z_{r} \cdot z_{p} / \tau )}{\sum_{a \in A(r)} \exp ( z_{r} \cdot z_{a} / \tau )},$$
where $r$ is the anchor (current) sample, and $R$ is the set of all samples ($X$ and $\tilde{X}$) in a training batch ($r \in R$); $P(r)$ is the set of samples that are positive pairs with anchor sample $r$ ($p \in P(r)$); $A(r)$ is the set of all samples in $R$ except for anchor sample $r$ ($a \in A(r) \equiv R \backslash \{r\}$); $z_{r}$, $z_{p}$ and $z_{a}$ are contrastive features obtained from $Proj\left(\cdot\right)$ for samples $r$, $p$ and $a$; and $\tau$ (temperature) is a tuneable hyperparameter for the contrastive loss.

\subsubsection{Contrastive Learning and Fine-tuning}
The overall process of contrastive learning and fine-tuning (Fig.~\ref{fig_scr}~(a)) is as follows. In contrastive representation learning, training samples (from $X$ and $\tilde{X}$) are input into the encoder $Enc\left(\cdot\right)$ and projector $Proj\left(\cdot\right)$ to get embeddings ($Z$ and $\tilde{Z}$). The supervised contrastive loss is computed using normalized embeddings ($Z$ and $\tilde{Z}$), where positive and negative pairs are determined by absolute differences between regression labels $Y$. After the contrastive representation learning, the parameters of $Enc\left(\cdot\right)$ are frozen and the $Proj\left(\cdot\right)$ is untouched, as in \citep{Chen2020-pz,Khosla2020-mx,Bahri2021-li,Xue2022-yz}. The usage of $Proj\left(\cdot\right)$ may retain useful information for downstream regression tasks in $Enc\left(\cdot\right)$ \citep{Chen2020-pz}. A predictor head for regression $Reg\left(\cdot\right)$ is added on top of the trained $Enc\left(\cdot\right)$. $Reg\left(\cdot\right)$ takes the output of $Enc\left(\cdot\right)$ as the input and is fine-tuned with MSE loss to obtain the final prediction.

\subsection{Ensemble Learning}
We use ensemble learning \citep{Hastie2009-uu} to combine prediction results from three predictors that are trained on three microstructural measures (FA, MD, and NoS) independently, as in \citep{He2022-po}. The ensemble prediction is obtained as the average prediction across the three predictors. Therefore, ensemble learning is beneficial in our application to study the relationship between three microstructural measures and neurocognitive performance metrics. Ensemble learning can also potentially improve the performance of the regression, because different microstructural measures may provide complementary information for prediction of neurocognitive performance. (Note that ensemble learning is used not only for our method but also for all compared methods in experiments.)

\subsection{Permutation Feature Importance}
We propose a permutation feature importance algorithm to assess the contribution of each cluster to the prediction of a neurocognitive score. Our proposed interpretation method is based on the permutation feature importance \citep{Breiman2001-hv}, which is a popular model-agnostic technique for estimating how important a feature is for a particular model. The traditional permutation feature importance is defined as the decrease in a model score (e.g., prediction accuracy) when a single feature value is randomly shuffled (permuted) across samples. This enables identification of highly important features that have a large effect on the model’s prediction accuracy. This traditional permutation feature importance method is not directly applicable to our high-dimensional data because the decrease of prediction accuracy is negligible when only permuting a single feature value. (Our input includes 953 cluster-specific white matter features per subject.) Therefore, we propose a new strategy to permute multiple feature values simultaneously (e.g., a random sample of 10\% of features). By repeating this strategy a very large number of times (e.g., 50,000), we can estimate the importance of all high-dimensional input features.

\subsection{Implementation Details}
For model training and performance evaluation, datasets are split into train/validation/test with the rate 70\%/10\%/20\%, and we repeat each experiment 10 times with different train/validation/test splits to report the average performance. Regarding the network structure, as suggested in \citep{Bahri2021-li}, $Enc\left(\cdot\right)$, $Proj\left(\cdot\right)$ and $Reg\left(\cdot\right)$ all have hidden dimension 256 with the ReLU activation in each layer. $Enc\left(\cdot\right)$ has four layers, whereas $Proj\left(\cdot\right)$ and $Reg\left(\cdot\right)$ both consist of two layers. For training hyperparameters, all deep learning methods are trained with the Adam optimizer with the learning rate 0.001 and use early stopping with patience 3 on the validation loss as in \citep{Bahri2021-li}. We conduct a grid search for parameter selection with $b \in \{256, 512, 1024, 2048, 4096\}$, $c \in \{0.3, 0.4, 0.5, 0.6, 0.7\}$, and $\tau \in \{0.5, 1, 5, 10\}$ for our method and all compared representation learning methods. For AdaCon, we also tune the temperature scaling factor ($ s\in \{10, 50, 100, 150\}$) based on their paper and code. Weight ratios of two losses in AdaCon are tuned with the rule that two losses should have similar values \citep{Dai2022-xu}. Then we choose batch size $b$ of 2048, corruption rate $c$ of 0.5, and temperature $\tau$ of 1 for our contrastive representation learning. Note that our method is not sensitive to hyperparameter changes and has good performance overall. Results with other parameter settings are presented in Section~\ref{subsubsec_hyperparameter} to demonstrate the robustness. A typical batch size of 128 is chosen in fine-tuning for all deep learning methods. Experiments are performed with Pytorch [16] (v1.8) on a NVIDIA GeForce RTX 2080 Ti GPU machine. For TractoSCR, each experiment (including training, validating and testing) takes about 30 seconds with 1.67GB GPU memory usage.

For the interpretation of prediction results, we implement our proposed feature permutation algorithm for prediction of three neurocognitive measures (PC1, General Ability; PC2, Executive Function; PC3, Learning/Memory) independently. For each permutation, we shuffle 95 out of 953 feature values across samples in the training dataset. Then we train using TractoSCR. The prediction accuracy is evaluated on the testing dataset, and the decrease of prediction accuracy (compared to the original prediction accuracy) is recorded along with the indices of the 95 shuffled features. For each of the 10 train/validation/test data distributions, we repeat this experiment 50,000 times (50,000 permutations). We obtain final overall importance scores for each feature (cluster) by averaging all recorded decreases of prediction accuracy from all permutations of that feature. Finally, three importance scores are obtained for each cluster, corresponding to the three prediction tasks. 

\section{Experiments and Results}
\label{sec_experiments}
\subsection{Evaluation Metric}
We computed Pearson correlation coefficients (Pearson's $r$) between the ground truth scores and predicted scores to quantify the prediction accuracy. The Pearson correlation coefficient is widely used for evaluation of cognitive prediction from neuroimaging data \citep{Cui2018-uu,Jollans2019-ma,Gong2021-up,Mansour_L2021-xs,Sripada2020-pg,Feng2022-xf,Chen2022-wq,Jandric2022-ki}. It measures the linear correlation (normalized cosine similarity) between two sets of data. A higher value of $r$ indicates a better prediction accuracy. We repeated each experiment 10 times with different train/validation/test splits (all methods use the same split). The mean and standard deviation of Pearson correlation coefficients across 10 splits are reported.

\subsection{Evaluation Results}
\subsubsection{Comparison of Representation Learning Methods}
We compared our proposed TractoSCR with one classical method (AutoEncoder \citep{Rumelhart1986-jc}), two recently proposed methods (VIME \citep{Yoon2020-xu} and SCARF \citep{Bahri2021-li}) for representation learning using tabular data, and one recent contrastive learning method (AdaCon \citep{Dai2022-xu}) for medical image-based regression. The autoencoder method is widely used for learning efficient representations. Here, the autoencoder has the same input as TractoSCR and the output has the same dimensionality as the input, and the MSE loss is applied. VIME uses a novel pretext task and data augmentation method for representation learning, and SCARF uses contrastive learning with random feature corruption. AdaCon utilizes its proposed contrastive loss together with an MSE loss for training, and for fair comparison to our method, we apply random corruption for data augmentation for AdaCon. In our study, we train these methods using the suggested settings in their papers and released codes.

\begin{table*}[h]
\caption{Comparison results (mean and standard deviation of Pearson's $r$ across splits) for prediction of three neurocognitive component scores, PC1 (General Ability), PC2 (Executive Function), and PC3 (Learning/Memory)}
\centering
\setlength{\tabcolsep}{2.5pt}  
\begin{tabular}{|c|l|c|c|c|}
\hline
 &  Methods & PC1 & PC2 & PC3 \\
\hline
\multirow{4}{*}{\thead{Representation \\ Learning \\ Comparison}} &  Autoencoder & 0.406±0.016 & 0.217±0.022 & 0.234±0.021\\
& VIME & 0.407±0.013 & 0.218±0.014 & 0.235±0.017\\
& SCARF & 0.411±0.013 & 0.217±0.019 & 0.239±0.020\\
& AdaCon & 0.414±0.013 & 0.235±0.021 & 0.253±0.020\\
\hline
\multirow{3}{*}{\thead{SOTA \\ Regression \\ Models}} &
ElasticNet & 0.400±0.018 & 0.206±0.021 & 0.237±0.016\\
& GBDT & 0.390±0.012 & 0.219±0.019 & 0.238±0.021\\
& MLP (baseline) & 0.409±0.018 & 0.209±0.020 & 0.236±0.020 \\
\hline
\multirow{2}{*}{\thead{Ablation \\ Study}} &  
TractoSCR\textsubscript{\textit{no-pd-fc}} & 0.407±0.016 & 0.210±0.019 & 0.231±0.013 \\
& TractoSCR\textsubscript{\textit{no-fc}} & 0.419±0.011 & 0.232±0.020 & 0.256±0.013 \\
\hline
\multicolumn{2}{|c|}{TractoSCR (Ours)} & \textbf{0.424±0.014} & \textbf{0.241±0.014} & \textbf{0.270±0.015} \\
\hline
\end{tabular}
\label{tab_results}
\end{table*}

Table~\ref{tab_results} shows that our proposed method outperforms all compared methods on the three prediction tasks. Our method and AdaCon perform better than other representation learning methods. This result demonstrates the effectiveness of utilizing the relationship between regression labels during contrastive learning. Furthermore, compared to AdaCon, the prediction accuracy of our method achieves relative improvements of 2.4\%, 2.6\% and 6.7\% on the prediction of three neurocognitive measures. This illustrates that using regression labels to enable positive and negative pair determination in contrastive learning can improve results on prediction of neurocognitive measures.

\subsubsection{Comparison of State-of-the-art Methods for Regression}
We also compared our proposed method with two SOTA machine learning methods for regression (ElasticNet \citep{Zou2005-oq} and GBDT \citep{Chen2016-em,Prokhorenkova2018-mr}). ElasticNet is popularly used in cognitive prediction \citep{Cui2018-uu,Gong2021-up}. It performs linear regression with L1 and L2 regularization. We used the implementation in the sklearn package \citep{Pedregosa2011-cx}. GBDT is a strong non-deep competitor for deep learning methods in tabular data \citep{Gorishniy2021-ll}. It iteratively constructs an ensemble of weak decision tree learners through boosting. We selected XGBoost \citep{Chen2016-em}, one of the most popular implementations of GBDT, for comparison. Parameters were tuned based on suggestions in \citep{Gorishniy2021-ll}. In addition to the above SOTA methods, we also included a multilayer perceptron (MLP) that has the same network structure as ours for a baseline comparison. As shown in Table~\ref{tab_results}, MLP (our baseline) outperforms ElasticNet and is competitive with GBDT. These results illustrate the power of deep learning methods for neurocognitive score prediction. In addition, 
compared to the MLP baseline, our proposed method obtains relative improvements in prediction accuracy of 3.7\%, 15.3\%, and 14.4\% on all three prediction tasks. This demonstrates the effectiveness of our proposed TractoSCR method. 

\subsubsection{Comparison of Ablated Versions}
An ablation study was conducted with two ablated versions (TractoSCR\textsubscript{\textit{no-pd-fc}} and TractoSCR\textsubscript{\textit{no-fc}}) of our proposed approach. TractoSCR\textsubscript{\textit{no-pd-fc}} performs contrastive learning without using regression labels for pair determination and without using random feature corruption. TractoSCR\textsubscript{\textit{no-fc}} uses regression labels for pair determination but does not perform random feature corruption. As shown in Table~\ref{tab_results}, the comparison between TractoSCR\textsubscript{\textit{no-pd-fc}} and TractoSCR\textsubscript{\textit{no-fc}} illustrates a large improvement when using regression labels for pair determination in contrastive learning. In addition, by applying random feature corruption for data augmentation, the performance improves on all tasks.

\subsubsection{Experiments under Different Hyperparameter Settings}
\label{subsubsec_hyperparameter}
\begin{figure}[!b]
\centerline{\includegraphics[width=\columnwidth]{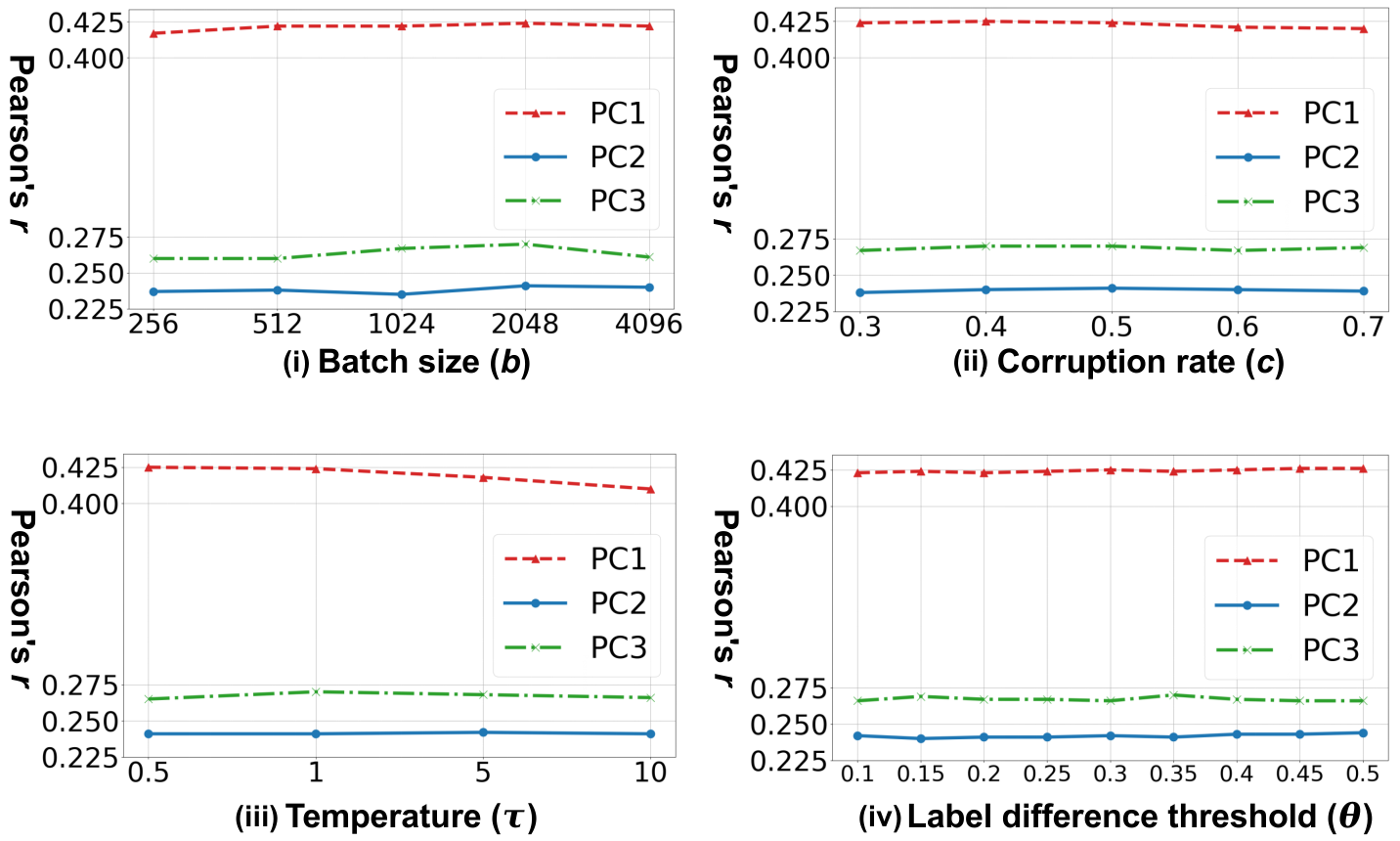}}
\caption{Hyperparameter sensitivity experiments for TractoSCR. Results (Pearson’s $r$) on predicting three neurocognitive component scores (PC1, PC2, and PC3) across different hyperparameters: (i) batch size $b$, (ii) corruption rate $c$, (iii) temperature $\tau$ , (iv) label difference threshold $\theta$. Results demonstrate that TractoSCR is \textit{hyperparameter-insensitive}.}
\label{fig_hyperparams}
\end{figure}

Fig.~\ref{fig_hyperparams} shows the accuracy of prediction of three neurocognitive component scores across four important hyperparameters in TractoSCR. Overall, TractoSCR achieves consistently high prediction accuracy (Pearson’s $r$) on all three tasks, which demonstrates TractoSCR is robust to hyperparameter change. Batch sizes and temperatures are important to contrastive learning frameworks in general \citep{Chen2020-pz,Khosla2020-mx}. Fig.~\ref{fig_hyperparams} (i) and (iii) show that TractoSCR obtains similar results when the batch size changes from 256 to 4096 and the temperature changes from 0.5 to 10. Corruption rates control how heavy the data augmentation is in contrastive learning \citep{Bahri2021-li,Yoon2020-xu}. A negligible change of the result occurs when corruption rates are varied from 0.3 to 0.7. The label difference threshold $\theta$ is the key parameter for positive and negative pair determination in TractoSCR. TractoSCR performs well under different $\theta$ thresholds ranging from 0.1 to 0.5. 

\begin{table}[!b]
\caption{Number of predictive fiber clusters within each anatomical category. Categories with the highest number of predictive clusters are in bold.}
\centering
\begin{tabular}{|l|l|l|l|}
\hline
 & PC1 & PC2 & PC3 \\
\hline
 Association & 8 (16.0\%) & 7 (14.0\%) & 13 (26.0\%) \\
 Projection & \textbf{15 (30.0\%)} & 9 (18.0\%) & \textbf{14 (28.0\%)} \\
 Commissural & 10 (20.0\%) & 2 (4.0\%) & 3 (6.0\%) \\
 Cerebellar & 5 (10.0\%) & 5 (10.0\%) & 6 (12.0\%) \\
 Superficial & 12 (24.0\%) & \textbf{27 (54.0\%)} & \textbf{14 (28.0\%)} \\
 Total & 50 (100.0\%) & 50 (100.0\%) & 50 (100.0\%) \\
\hline
\end{tabular}
\label{tab_anatomical}
\end{table}

\subsection{Interpretation Results}
\begin{figure*}[!t]
\centerline{\includegraphics[width=\textwidth]{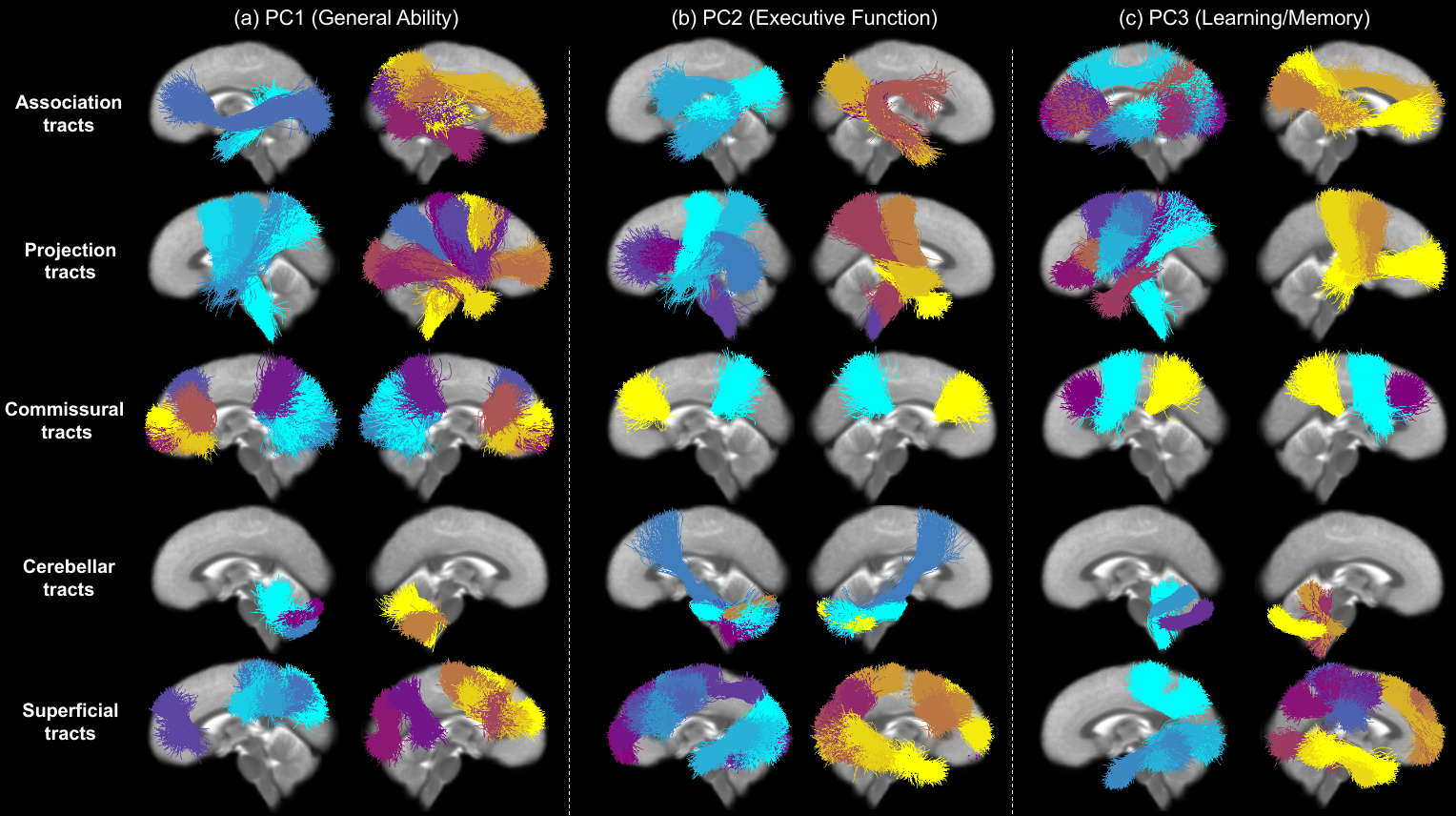}}
\caption{Visual presentation of most predictive fiber clusters (with the 50 highest importance scores) for each individual prediction task. Different fiber clusters are depicted in different colors and organized according to five anatomical tract categories. }
\label{fig_visual_interpret}
\end{figure*}

Fig.~\ref{fig_visual_interpret} provides a visualization of the most predictive fiber clusters (defined as the fiber clusters with the top 50 highest importance scores for each prediction task). Together, these fiber clusters may form part of the putative structural networks relating to general cognitive ability (PC1), executive function (PC2), and learning/memory (PC3). The predictive fiber clusters span across all five anatomical tract categories (association, cerebellar, commissural, projection, and superficial tracts) \citep{Zhang2018-jx} and are found in both the left and right hemispheres. This finding is in line with neurocognitive research demonstrating that higher order cognitive functions, such as the ones presently under investigation, are broadly distributed across the brain \citep{Goddings2021-ip}. When this result is examined in detail, we find that the predictive fiber clusters are predominantly located within the superficial and projection white matter (Table~\ref{tab_anatomical}). This finding contrasts with the relative plethora of white matter and cognition studies that have focused on the role of the association connections (e.g., language in arcuate fasciculus, memory in the uncinate fasciculus, etc.) \citep{Forkel2022-il}. Details about the location of all predictive fiber clusters (Fig.~\ref{fig_visual_interpret}) within specific tracts (as defined in the anatomically curated ORG atlas \citep{Zhang2018-jx}) are provided in Supplementary Table S1. Overall, the most predictive tracts are the superficial frontal white matter and striato-frontal connections, which have the highest number of clusters found to be important across the three prediction tasks.

\section{Discussion}
\label{sec_discussion}

In this study, we proposed a novel deep-learning-based regression method that enables improved prediction accuracy of neurocognitive measures. To our knowledge, we are the first to focus on deep representation learning for neuroimage-based prediction of neurocognitive measures. Unlike commonly used regression methods \citep{noauthor_undated-tc},\citep{Feng2022-xf},\citep{Madole2021-io},\citep{Li2020-le}, the proposed TractoSCR method allows us to effectively leverage information from regression labels during contrastive learning. A new strategy was proposed to use the absolute difference between two continuous regression labels to determine positive and negative pairs. We also employed random feature corruption, a data augmentation method for tabular data, in contrastive learning. By applying random feature corruption, the performance improved on all prediction tasks (e.g., a relative improvement of 5.5\% on PC3). 

We showed that our proposed method achieved highly improved prediction performance on a large-scale ABCD dataset compared to existing methods, including SOTA regression methods and representation learning methods. For example, on PC3, our method outperformed the SOTA contrastive learning method (AdaCon) with a relative improvement of 6.7\% in Pearson’s $r$, and our method outperformed the baseline method (MLP) with a relative improvement of 14.4\% in Pearson’s $r$. We also illustrated that TractoSCR is robust to changes of hyperparameters (batch size $b$, corruption rate $c$, temperature $\tau$, and label difference threshold $\theta$). These results demonstrate the utility of contrastive representation learning methods for the neuroimaging-based prediction of higher-order cognitive abilities. In this study, we obtained Pearson's $r$ values ranging from 0.24 to 0.43, indicating a moderate correlation between investigated white matter microstructural measures and neurocognitive scores. Our moderate correlation finding is in general in line with a body of recent work that uses neuroimaging measures to predict cognition~\citep{Sripada2020-pg, Gong2021-up, Kim2021-kp, Feng2022-xf}.

Predicting neurocognitive measures from the ABCD dataset is an interesting but challenging task that has been undertaken using various MRI modalities \citep{Kilian_M_Pohl_Wesley_K_Thompson_Ehsan_Adeli_Marius_George_Linguraru_undated-nh,Sripada2020-pg,Ooi2022-qt}. For example, T1-weighted MRI was used to predict fluid intelligence scores \citep{Kilian_M_Pohl_Wesley_K_Thompson_Ehsan_Adeli_Marius_George_Linguraru_undated-nh}, while a comparison across modalities suggested that information from fMRI could best predict a summary cognition score derived from 36 behavioral scores \citep{Ooi2022-qt}. One recent study by Sripada et al. \citep{Sripada2020-pg} used resting-state fMRI to predict the same neurocognitive component scores (PC1, PC2, and PC3) that we have investigated in the current study. Their method obtained Pearson's $r$ values of 0.33, 0.09, and 0.15 for the prediction of PC1, PC2, and PC3, respectively \citep{Sripada2020-pg}. These results were based on a smaller dataset (2013 subjects from the first ABCD data release) and are not directly comparable to our results. However, we note that using tractography fiber cluster microstructure features as input and our novel TractoSCR regression framework for prediction, we obtained higher Pearson's $r$ coefficients of 0.42, 0.24, and 0.27 for the prediction of PC1, PC2, and PC3, respectively. Overall, this suggests that fiber cluster measures can potentially provide highly informative features, in combination with TractoSCR that achieves higher prediction accuracy than commonly used linear regression methods.

In our data-driven analysis of imaging and neurocognitive data from 8735 participants of the ABCD study, we found that fiber clusters within the projection and superficial white matter were the most important for predicting neurocognitive scores related to general cognitive ability, executive function, and learning/memory. This result was enabled by the proposed permutation feature importance algorithm for identifying predictive features from high-dimensional input. This finding may highlight the need for more investigations of the superficial and projection pathways in the context of cognition.

Potential limitations and future work of the present study are as follows. First, in the present study, we explored the relationships between neurocognitive scores and fiber cluster microstructural measures from a single imaging modality, dMRI. Future work may investigate TractoSCR for predicting neurocognitive scores based on features from multiple MRI modalities. Second, we focused on prediction of neurocognitive scores in healthy children. Future work may investigate the proposed TractoSCR framework to predict cognition in the context of aging or disease (e.g., Alzheimer’s Disease \citep{Fisher2019-zg}). Third, we employed a relatively simple MLP network. Future developments can include the incorporation of more advanced  deep learning networks (e.g., transformer \citep{Vaswani2017-fc}) and recently proposed regression losses \citep{Engilberge2019-ee,Li2020-bz,Chen2022-ma}. Finally, our results demonstrate the utility of contrastive representation learning for neuroimaging-based prediction of cognition. However, our proposed TractoSCR and permutation feature importance methods can be applied to other regression tasks, and assessment of their performance is left for future work.

\section{Conclusion}
\label{sec_conclusion}
In this work, we have proposed TractoSCR, a simple yet effective contrastive representation learning method for regression. We applied our TractoSCR method on multi-site harmonized dMRI tractography measures from the large-scale ABCD dataset (8735 participants) to predict neurocognitive scores relating to general cognitive ability, executive function and learning/memory. We compared TractoSCR with several SOTA methods and showed highly improved prediction performance. Overall, we found that fiber clusters within the projection and superficial white matter were the most important for predicting neurocognitive scores.

\section*{Human Ethics Statements}
Approval was granted by the BWH IRB for use of the public ABCD data.

\section*{Data and Code Availability}
ABCD data used in this study can be downloaded at https://nda.nih.gov/abcd. Code will be made available.

\section*{Acknowledgments}
We acknowledge the following NIH grants: P41EB015902, R01MH074794, R01MH125860, R01NS125781, R01NS125307, and R01MH119222. F.Z. also acknowledges a BWH Radiology Research Pilot Grant Award.



 \bibliographystyle{elsarticle-harv} 
 \bibliography{refs}





\end{document}